# A Self-Attention Joint Model for Spoken Language Understanding in Situational Dialog Applications

Mengyang Chen[1], Jin Zeng[2] and Jie Lou[3]
[1]ByteDance Corporation, China, chenmengyang@bytedance.com
[2] ByteDance Corporation, China, zengjin@ bytedance.com
[3] ByteDance Corporation, China, loujie.roger@ bytedance.com

## I. INTRODUCTION

Spoken language understanding (SLU) acts as a critical component in goal-oriented dialog systems. It typically involves identifying the speaker's intent and extracting semantic slots from user utterances, which are known as intent detection (ID) and slot filling (SF). An example from airline travel information system (ATIS) [1] corpus is demonstrated in Table I. SLU problem has been intensively investigated in recent years. However, these methods just constrain SF results grammatically, solve ID and SF independently, or do not fully utilize the mutual impact of the two tasks. This paper proposes a multi-head self-attention joint model with a conditional random field (CRF) layer and a prior mask. The experiments show the effectiveness of our model, as compared with state-of-the-art models. Meanwhile, online education in China has made great progress in the last few years. But there are few intelligent educational dialog applications for students to learn foreign languages. Hence, we design an intelligent dialog robot equipped with different scenario settings to help students learn communication skills.

| sentence | slots | intent |
|---|---|---|
| All | O | |
| flights | O | |
| from | O | flight |
| boston | B-fromloc.city_name | |
| to | O | |
| washington | B-toloc.city_name | |

TABLE I. AN EXAMPLE OF SLOTS AND INTENT LABEL IN ATIS

## II. RELEVANT THEORIES OF LEARNING

### A. Intent Detection

Intent detection (ID) is a standard classification problem. One intent label is predicted for a sentence. Traditional classifiers such as support vector machine (SVM) can be applied [2]. Also, neural network methods can be utilized through extracting text features by convolutions [3].

### B. Slot Filling

Slot filling (SF) is formulated as a sequence labeling task. A slot label is predicted for each word in the sentence. Classical model, such as conditional random field (CRF) is employed in [4]. Recently, the recurrent neural network (RNN) is introduced to solve language understanding, due to its ability to capture the dependency between words. [6] adopts Long short-term memory (LSTM) [5] to achieve slot labels. Modifications based on [6] have been proposed, such as adding a regression layer in [7] or a CRF layer in [8] upon the LSTM layer to capture the relation between tags. In addition, [9] employs a bidirectional-LSTM for char embedding in case of new words occurrence.

Considering that ID and SF have semantic relation, a number of studies assemble these two tasks in order to achieve mutual promotion. [10] firstly introduces a joint model which utilizes a shared CNN to extracted features for ID and SF and a triangular CRF to capture their relation. [11] exploits shared context embedding and LSTM instead of CNN. [12] attempts to employ the attention-based seq2seq framework. [13] uses a CRF layer to constrain LSTM output but do not explore the potential impact of the outputs, while [14] applies bidirectional-LSTM with self-attention and gate mechanism to fuse IC and SF outputs. [15] proposes a similar work as [14] but uses more complicated attention mechanism and achieves the state-of-the-art result in ATIS currently.

However, none of the above methods deal with (i) constraining SF output grammatically (ii) joint modeling on ID and SF (iii) fully exploring the correlation between slots and intent outputs simultaneously. Therefore, we propose a joint model, which uses multi-head local self-attention to extract shared features, a mask gating mechanism to explore the correlation of outputs, and a CRF to constrain the SF output, to solve the problem perfectly.

## III. ENABLING TECHNOLOGICAL ADVANCES

### A. Framework

We present our model in this section and an overview is shown in Figure 1. The first layer maps input sequence $X = \{x_k\}$ into vectors by concatenating its word-level embeddings $\{e_k^w\}$ and character-level embeddings $\{e_k^c\}$ obtained from bidirectional-LSTM [10], where $k$ is the index of words in sequence. Since the contextual information, especially the adjacent words, are useful in sequence labeling, we then employ a multi-head local self-attention to extract context-aware features $\{c_k^e\}$. The local context features are $H = (h_{k-w}, \dots, h_k, \dots, h_{k+w})$ and attention output is calculated as:

$$a = softmax(W_{km1} \tanh(W_{km2} H)) \quad (1)$$
$$c_k^e = aH \quad (2)$$

where $W_{km1}$ and $W_{km2}$ are the first and second weight matrixes of local self-attention for $k^{th}$ word and $m^{th}$ head. The attention width we adopt is 5 and the head number is 2. We then introduce a bidirectional-LSTM layer to capture the dependency and produce $\{h_k\}$, which is applied to classify the intent $y^I$ using a multi-layer fully-connected classifier.

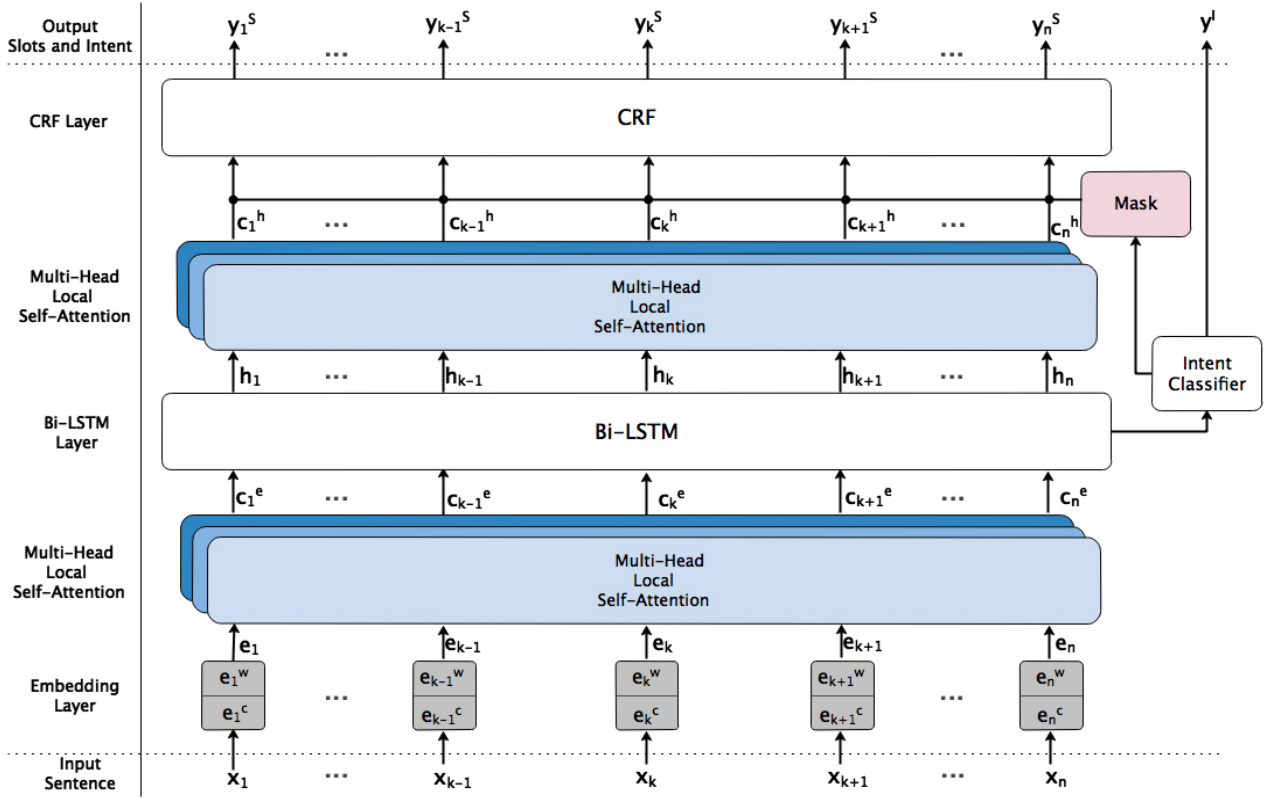

Fig. 1. Framework of intent and slot filling joint model

$$y^I = softmax(W^I h_n + b^I) \tag{3}$$

A cross-entropy loss is used during training and the intent label corresponds to the index that gives the highest probability during prediction.

The hidden states are also sent to a similar multi-head local self-attention structure to generate $\{c_k^h\}$. Since the distributions of slots under different intents are quite different, we adopt a prior mask, which is actually a conditional probability distribution of slot given intent $P(y_k^S|y^I)$, obtained from training data. We multiply the mask and intent output, concatenate the result and $\{c_k^h\}$, finally put it into a CRF layer to get grammatically constrained slot results $Y^S = \{y_k^S\}$. We consider P to be the matrix of scores output by attention. $P_{k,y_k^S}$ represents the score of $y_k^S$ label of $k^{th}$ word. We define the score function to be:

$$s(X, Y^S) = \sum_{k=0}^{n} A_{y_k^S, y_{k+1}^S} + \sum_{k=1}^{n} P_{k, y_k^S} \tag{4}$$

A softmax over all possible tag sequences yields a probability for the sequence $Y^S$. We maximize the log-probability of the correct tag sequence during training. While decoding, we predict the output sequence with the maximum score.

### B. Experiments

In order to evaluate the efficiency of the proposed model, we conduct experiments on ATIS and Snips datasets, which are widely used as benchmarks in SLU research. ATIS contains audio recordings of people making flight reservations. Snips is collected from personal voice assistants. The performance of the slot filling task is measured by the F1-score, while intent detection task is evaluated with prediction accuracy. Results of our model against other methods are listed in Table II. Our method shows 0.14% and 0.49% improvements in ID on the two datasets, 0.02% and 0.04% in SF, as compared with state-of-the-art methods.

| Model | ATIS | | Snips | |
|---|---|---|---|---|
| | Slot (F1) | Intent (Acc) | Slot (F1) | Intent (Acc) |
| Atten-Based [12] | 95.98 | 98.43 | 87.8 | 96.7 |
| Slot-Gated [14] | 95.20 | 94.10 | 88.80 | 97.00 |
| BLSTM-CRF [13] | 95.62 | 97.42 | **93.90** | 99.22 |
| Self-Attention [15] | **96.52** | 98.77 | -- | -- |
| **Our Model** | **96.54** | **98.91** | **93.94** | **99.71** |

TABLE II. SLU PERFORMANCE ON ATIS AND SNIPS DATASETS (%)

## IV. REAL WORLD APPLICATIONS

For a long time in the past, non-native English students spent too much time on grammar or reading comprehension in order to cope with examinations, resulting in lack of listening and speaking ability. Recently, many applications, such as VIPKID and English Liulishuo, appear on the software market. However, most of the them either focus on speech shadowing, or simple conversations without complicated dialog state tracking, thus cannot help students express fluently and deal with conversations under complicated scenes.

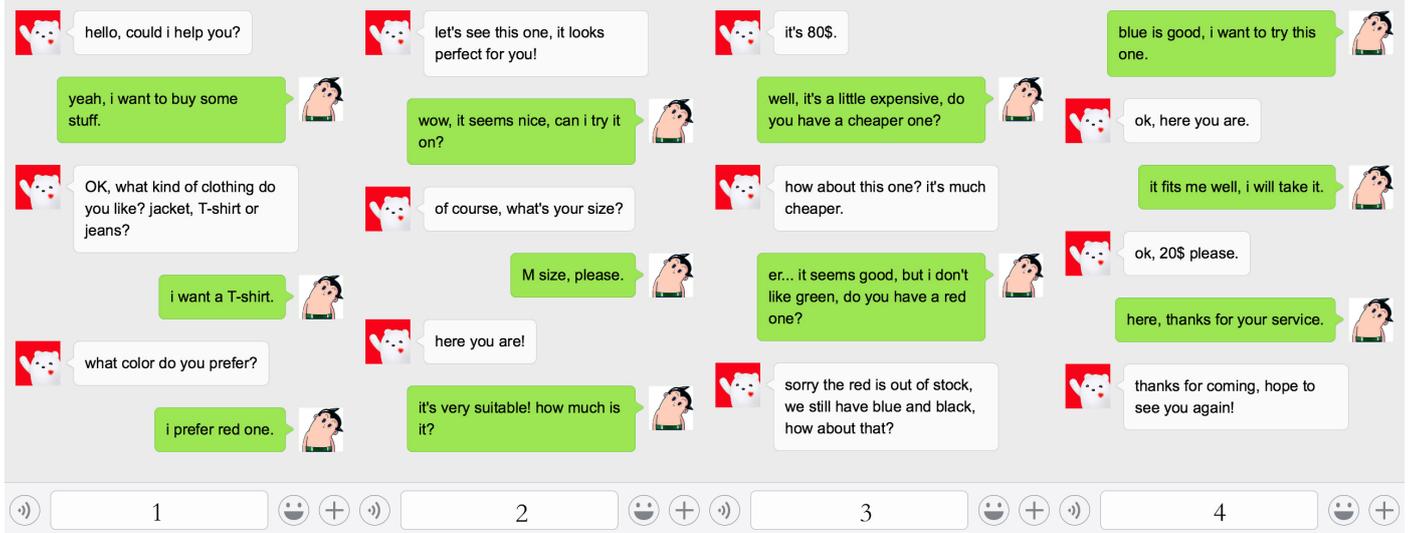

Fig. 2. An example of shopping in our APP

To help K12 students better deal with realistic scenarios such as traveling, shopping, and restaurant ordering, we design a conversational dialog APP. The framework is actually a standard goal-oriented dialog system and is illustrated in Figure 4. Firstly, the agent adopts the previously discussed SLU method to analyze students' intent and slot-values. Secondly, it uses rule-based dialog tracking skills to record conversation state changes and make relevant actions. At last, it uses pre-defined templates to generate responses. Since the topic discussed in this paper is SLU, we remove the ASR and TTS modules and use a simplified version for convenience (we adopt BAIDU speech API in practice). Figure 2 shows a shopping example and Figure 3 is the pseudocode of DST and logic of decision-making.

Students can get tips by selecting "help" in APP when they do not know how to respond.

## V. EVIDENCE OF POTENTIAL IMPACTS

Education is a field that people all around the world attach great importance to. Related researches show that there are only a quarter of students in China could enter undergraduate studies, which is far fewer than developed countries. Fortunately, the education industry has made great progress in the last few years. The growth rate of K12 market size in China maintains at more than 30% since 2013. However, traditional companies encounter profit issues because of huge costs of rent and teacher salary. They turn their focal point to online education schema, in which the CR4 (Four-firm Concentration Rate) is under 5% and no giant company exists currently. Moreover, China has made government plans to encourage burgeoning market forces to enter this field. As for parents, they highly value English learning and are willing to pay for strengthening their kids' speaking and listening skills. Nevertheless, popular applications, such as VIPKID, focus on speech shadowing, and cannot fully meet their needs. As a matter of fact, the core objective of

---

**Algorithm 1** Agent Dialog State Tracking and Decision Making Logic

$k$: the turn index
$U_k$: user input utterance
$R_k$: agent response
$I_k$: detected intent or user action
$C_{ik}$: detected slot-value chunks
$A_k$: agent action
$\{RS_m\}$: request slot-values
$\{IS_n\}$: inform slot-values
$\{DS_p\}$: deny slot-values
**while True do**
    user input $U_k$
    SLU detect $A_k, C_{ik}$
    **if** $I_k ==$ byemsg **then**
        break
    **else if** $I_k ==$ greeting **then**
        $A_k =$ greeting
    **else if** $I_k ==$ inform **then**
        $A_k =$ random_choice(inform, reqeust); tracker update $IS$
    **else if** $I_k ==$ request **then**
        $A_k =$ inform; tracker update $RS$; agent search database
    **else if** $I_k ==$ ask_recommend **then**
        $A_k =$ recommend; agent search database
    **else if** $I_k ==$ deny **then**
        $A_k =$ random_choice(reqeust, recommend); tracker update $DS$
    **else**
        $A_k =$ tips (help chat continue)
    **end if**
    $R_k =$ template NLG using $A_k$ and formated $RS, IS, DS$
**end while**

Fig. 3. Pseudocode of dialog state tracking and decision-making logic

learning foreign languages is to speak and listen fluently in daily conversations. There are only a few companies concentrating on situational dialogues and they just use simple logic to track dialog states. In other words, the dialog cannot carry on without predefined user utterance. Students may be bored with such patterns. The designed conversational robot based on our method provides a greater degree of talking freedom and tracks dialog state transition among different intents and slot values. We conduct a user study with 50 K12 students sampled from local primary and middle schools in Beijing. Most of them consider this pattern fresh and show great enthusiasm to continue talking with the robot. The APP is now under internal test and will be released later.

## VI. Summary

In this paper, we propose a joint learning model for the SLU task. Local self-attention and embedding are performed to extract sentence features, which are sent to bidirectional-LSTM to capture the relation between words. Then, intent classification and slot filling tasks are conducted in a mask gating mechanism using the shared features. Moreover, we apply a CRF layer to constrain the output of slots and obtain a reasonable result. We also design an educational APP based on the proposed SLU method, rule-based tracking skills, and template-based language generation skills, to help students speak and listen well in practice. Furthermore, we establish a dataset of several common scene conversations. We will continue to enlarge the dataset and release it later.

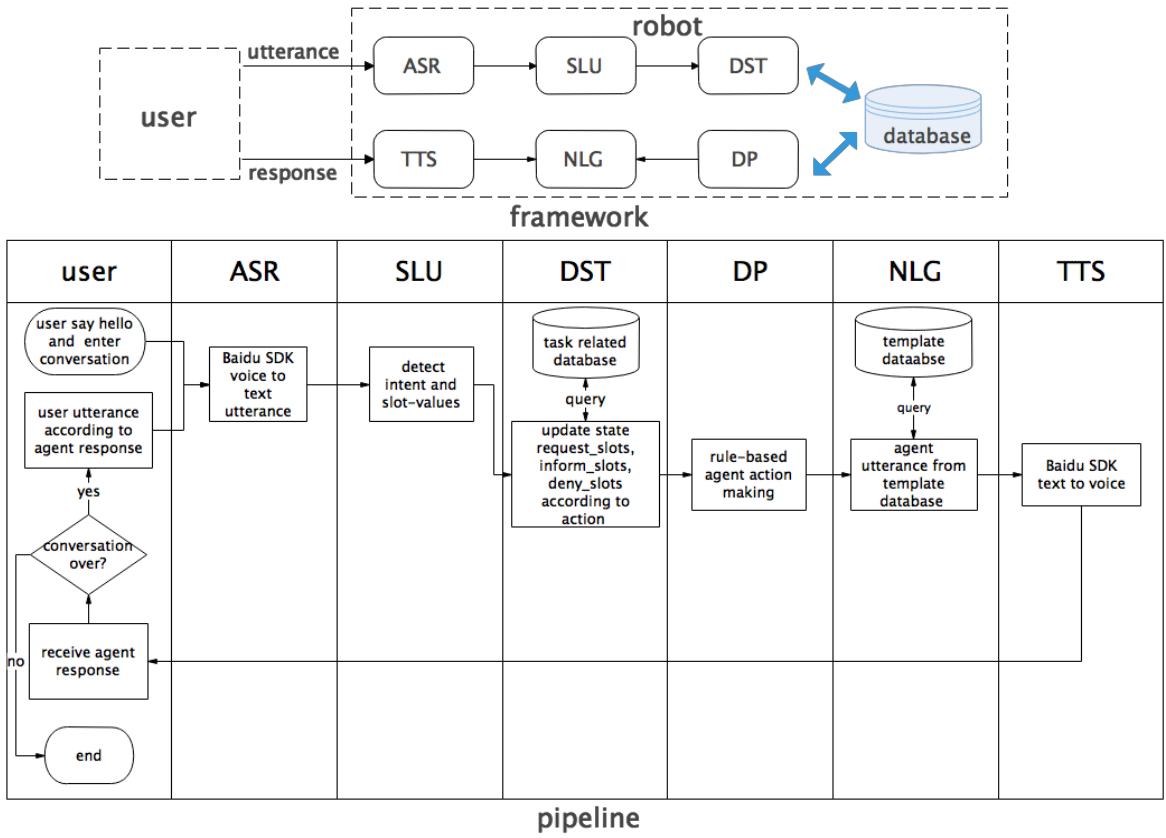

Fig. 4. Framework and pipeline of the situational dialog application